# PDCFNet: Enhancing Underwater Images through Pixel Difference Convolution and Cross-Level Feature Fusion

**Song Zhang** [a, b, c, d], **Daoliang Li** [a, b, c, d], **Ran Zhao** [a, b, c, d *]

B20203080601@cau.edu.cn; dliangl@cau.edu.cn; ran.zhao@cau.edu.cn

[a] National Innovation Center for Digital Fishery, China Agricultural University, Beijing 10083, China

[b] Key Laboratory of Smart Farming Technologies for Aquatic Animal and Livestock, Ministry of Agriculture and Rural Affairs, China Agricultural University, Beijing 10083, China

[c] Beijing Engineering and Technology Research Center for Internet of Things in Agriculture, China Agricultural University, Beijing 10083, China

[d] College of Information and Electrical Engineering, China Agricultural University, China Agricultural University, Beijing 10083, China

* Corresponding author at: P. O. Box 121, China Agricultural University, 17 Tsinghua East Road, Beijing 100083, China. E-mail address: ran.zhao@cau.edu.cn (R. Zhao)

**Abstract:** Majority of deep learning methods utilize vanilla convolution for enhancing underwater images. While vanilla convolution excels in capturing local features and learning the spatial hierarchical structure of images, it tends to smooth input images, which can somewhat limit feature expression and modeling. A prominent characteristic of underwater degraded images is blur, and the goal of enhancement is to make the textures and details (high-frequency features) in the images more visible. Therefore, we believe that leveraging high-frequency features can improve enhancement performance. To address this, we introduce Pixel Difference Convolution (PDC), which focuses on gradient information with significant changes in the image, thereby improving the modeling of enhanced images. We propose an underwater image enhancement network, PDCFNet, based on PDC and cross-level feature fusion. Specifically, we design a detail enhancement module based on PDC that employs parallel PDCs to capture high-frequency features, leading to better detail and texture enhancement. The designed cross-level feature fusion module performs operations such as concatenation and multiplication on features from different levels, ensuring sufficient interaction and enhancement between diverse features. Our proposed PDCFNet achieves a PSNR of 27.37 and an SSIM of 92.02 on the UIEB dataset, attaining the best performance to date. Our code is available at https://github.com/zhangsong1213/PDCFNet.

# 1. Introduction

As an important medium for information acquisition, transmission, and processing, underwater images play a crucial role in underwater operations. With the continuous deepening of exploration in the marine field, an increasing number of underwater tasks, such as underwater navigation [1], resource exploration [2], environmental monitoring [3], and fisheries operations [4], are becoming increasingly dependent on visual imaging to obtain more information. However, due to the unique underwater environment, light inevitably undergoes absorption and scattering[5], resulting in image degradation, such as low contrast, blurriness, and color distortion [6]. These degradations pose significant challenges to vision-based tasks. Enhancing the visibility of underwater images and improving the detectability of targets will undoubtedly ease the burden on downstream visual tasks. Therefore, utilizing underwater image enhancement (UIE) techniques to restore and improve the visual quality of underwater images is of critical importance.

UIE is a classic and challenging task in computer vision. Its goal is to restore degraded underwater images to their original appearance, enhancing image clarity, contrast, and color fidelity. Due to the unavailability of real ground truth for underwater scenes, underwater image enhancement and restoration become an ill-posed problem [6, 7], making them highly challenging. Physics-based methods aim to reverse the physical processes of optical scattering and absorption, removing blurring and distortion in images to restore a representation closer to the real scene [8-10]. However, physics-based methods typically rely on a single model, which struggles to describe the imaging process in complex scenes. Moreover, the prior information used for parameter estimation often fails in challenging situations. Non-physics-based methods enhance image clarity and detail by directly adjusting image contrast, brightness, and color balance [11-13]. However, due to a lack of robustness, non-physics-based methods may lead to over-enhancement or insufficient enhancement of underwater images.

In recent years, with the development of neural networks and their powerful feature representation capabilities, as well as the availability of large quantities of datasets, researchers have begun to leverage Convolutional Neural Networks (CNNs) to learn broader feature mappings from extensive datasets [14-16]. Compared to traditional physics-based and non-physics-based methods, deep learning-based approaches demonstrate impressive performance and robustness [17]. However, vast majority of UIE methods use vanilla convolution for modeling without incorporating any prior knowledge. While vanilla convolution excels at capturing local features and learning the spatial hierarchical structure of images, it is typically less sensitive to high-frequency details. This is primarily because vanilla convolution operations tend to smooth input images, leading to some degree of detail loss. Without proper constraints or specifically designed mechanisms to preserve or enhance high-frequency features, traditional CNN models may prioritize learning more prominent low-frequency features within their vast solution space, while neglecting subtle high-frequency information. This limitation restricts feature expression and modeling. A prominent characteristic of underwater degraded images is blur, and the goal of enhancement is to make textures and details (high-frequency features) more visible. Therefore, we believe that leveraging high-frequency features can effectively enhance underwater images.

Difference convolution is a convolution operation that captures local changes or gradient information by calculating the differences between adjacent pixels or features. Unlike vanilla convolution, which focuses on weighted summation within a certain receptive field to achieve a holistic representation of features, difference convolution emphasizes the disparities between local regions, enabling better extraction of image edges and details [18, 19]. This provides new insights and methods for addressing UIE task. To tackle above issues, we propose a single underwater image enhancement network, PDCFNet, based on Pixel Difference Convolution (PDC) and cross-level feature fusion. Specifically, we design a Detail Enhancement Module (DEM) based on PDC that employs parallel PDCs to capture high-frequency features, resulting in improved detail and texture enhancement. The designed Feature Fusion Module (FFM) performs operations such as concatenation and multiplication on features from different levels, ensuring sufficient interaction and enhancement among diverse features. We conducted comprehensive quantitative and qualitative analyses on three public datasets, demonstrating the effectiveness of the proposed PDCFNet. The main contributions of this paper are as follows:

1) We designed a novel UIE network, PDCFNet, based on PDC and cross-level feature fusion. This network includes a DEM to capture high-frequency features for better detail recovery, as well as a FFM to facilitate interaction and enhancement among features at different levels.
2) To the best of our knowledge, this is the first time that difference convolution has been introduced to UIE task. Specifically, we designed a

DEM that incorporates parallel convolution and PDCs, leveraging PDC to obtain richer textures and detail features.

3) Extensive qualitative and quantitative experiments demonstrate that our proposed PDCFNet outperforms current state-of-the-art models, providing a powerful and promising solution for UIE.

## 2. Related works

### 2.1. Deep learning-based UIE

Deep learning is a data-driven approach that automatically learns features from vast amounts of data. In recent years, deep learning-based UIE has progressed rapidly. Due to the lack of ground truth, Li et al. [20] proposed an unsupervised Generative Adversarial Network (GAN), WaterGAN, which synthesized degraded underwater images from air images and depth maps, followed by training on the synthesized paired dataset. Guo et al. [21] introduced a multi-scale densely connected GAN, enhancing performance by incorporating residual dense connection blocks in the generator and using spectral normalization in the discriminator to stabilize training. Inspired by fusion ideas [22], Li et al. [14] input preprocessed images alongside original images into a CNN to capture richer image information, proposing a multi-input single-output UIE network, WaterNet. Considering the lack of prior information guidance during end-to-end training, Wu et al. [23] decomposed underwater images into high-frequency and low-frequency components, introducing transmission maps and background light in the low-frequency component. Liu et al. [24] embedded an improved underwater image formation model [25] into their network design, employing GAN to estimate key parameters in the model, guiding the generation of high-quality enhanced images. Li et al. [17] and Zhang et al. [26] incorporated medium transmission maps and reflection maps as guiding information into their networks to facilitate the generation of high-quality underwater images. Cong et al. [27] designed a network to learn parameters for the inversion of physical models while proposing dual discriminators for content and style.

Some studies approach the problem from different perspectives. Fu et al. [16] examine UIE from a probabilistic standpoint, categorizing it into distribution estimation and consensus processes, and use conditional variational autoencoders along with a consensus process to predict deterministic outcomes. Sun et al. [28] propose a reinforcement learning-based framework, modeling the UIE as a Markov decision process. Chen et al. [29] introduce a content-style separation framework for domain adaptation, generating more natural underwater images by incorporating air images. Zhou et al. [30] propose an enhancement network that utilizes cross-view neighboring features, employing efficient feature alignment for adjacent features. Additionally, some studies focus on global modeling of UIE, introducing various Transformers [31-34].

**2.2. Difference convolution**

Difference convolution can be traced back to Local Binary Patterns (LBP) [35]. LBP assigns values (0 or 1) to neighborhoods by comparing the grayscale values of a pixel with those of its neighbors, forming binary patterns. LBP is computed based on predefined patterns and rules, with no learnable parameters [36]. In contrast, difference convolution calculates the differences between adjacent pixels using convolution kernels, producing continuous values rather than binary ones. Moreover, difference convolution is a learnable convolution operation that can be adapted to different task requirements by adjusting the weights of the convolution kernels. As a result, difference convolution offers superior expressive capability and adaptability in complex tasks [37, 38].

Due to the success of CNNs in computer vision tasks, Juefei-Xu et al. [39] proposed a Local Binary Convolution, which approximates the corresponding activation responses of standard convolutional layers by utilizing linear weights combined with activated filter responses. Su et al. [40] combined traditional edge detection operators with CNNs to propose Pixel Difference Convolution (PDC). Yu et al. [41] introduced Central Difference Convolution (CDC), which aggregates intensity and gradient information to capture intrinsic detailed patterns for facial anti-fraud. Zhang et al. [19] proposed a new difference convolution that extracts the pattern direction of a pixel by calculating the directional variation between a pixel and its adjacent pixels. Chen et al. [42] developed a detail enhancement convolution based on difference convolution for single-image dehazing. Given the ability of difference convolution to capture gradient information, we are the first to introduce difference convolution into the UIE task.

**3. Proposed method**

**3.1. Overall structure**

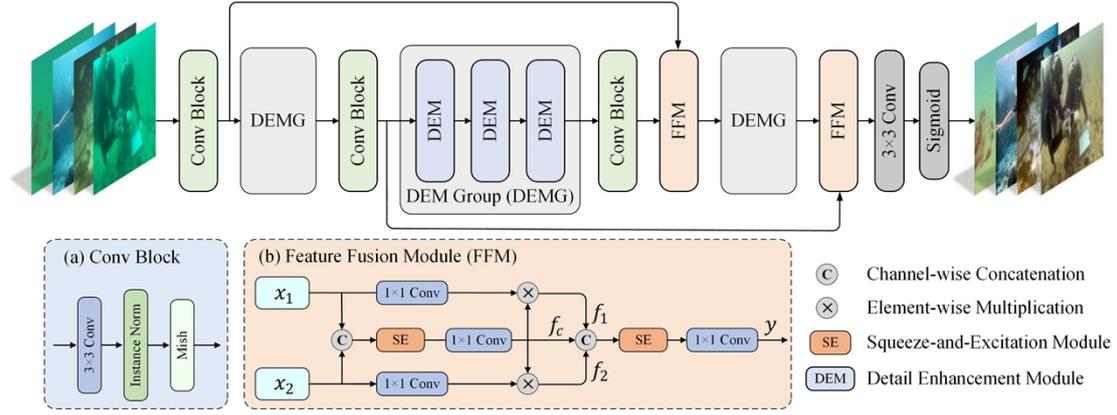

**Fig. 1.** Overall architecture of the proposed PDCFNet network.

Our proposed PDCFNet aims to address degradation issues such as blurriness and color distortion in underwater images through cross-level feature extraction and fusion, as well as DEMs, to achieve high-quality underwater image enhancement. As shown in Fig. 2, PDCFNet first employs a convolutional block for initial feature extraction from the input image, providing a foundation for subsequent detail enhancement and feature fusion. This convolutional block consists of a 3×3 convolution, instance normalization (IN), and a Mish activation function. IN calculates the mean and variance for each channel of every instance (i.e., each image) and performs normalization, which helps maintain the independence of the images and ensures stable training. Following this is a Detail Enhancement Module Group (DEMG), which contains three sets of DEMs. DEM utilizes PDC to extract rich image textures and detail features, aiding in the restoration and deblurring of underwater images. FFM performs operations such as concatenation and multiplication on features from different levels, ensuring sufficient interaction and enhancement among diverse features. PDCFNet employs two sets of FFMs and adopts a dual connection approach to fully utilize features from different levels. The output is produced using a 3×3 convolution followed by a Sigmoid function, mapping the results to the range of 0-255 to obtain the final enhanced image. Moreover, our proposed PDCFNet is a non-degradation network (i.e., it does not use downsampling operations), which effectively preserves detail information during the enhancement process.

### 3.2. Detail enhancement module

PDC captures image textures and details by emphasizing high-frequency features. The computation process of PDC is similar to that of vanilla convolution, where the pixel values in the feature map patch covered by the convolution kernel are

replaced by pixel differences. PDC can be described as follows:

$$y = f(\nabla x, \theta) = \sum_{(x_i, x_i') \in \mathcal{P}} w_i \cdot (x_i - x_i') \tag{1}$$

where $x_i$ and $x_i'$ are the input pixels, $w_i$ is the convolution kernel of size $k \times k$, and $\mathcal{P} = \{(x_1, x_1'), (x_2, x_2'), ...., (x_m, x_m')\}$ is the set of pixel pairs selected from the current patch, with $m \leq k \times k$.

To obtain rich gradient information, PDC employs LBP and its robust variant, Extended LBP (ELBP) [43-45], to encode pixel relationships from different directions (angles and radial positions). Specifically, ELBP first computes the pixel differences within a local patch to generate a pixel difference vector, which is then binarized to obtain a binary code of length $m$. This code distribution (or histogram) is calculated using a bag-of-words model [46] as the image representation. ELBP is then integrated with CNN convolutions, resulting in three instances of PDC, as illustrated in Fig. 2, referred to as Central Pixel Difference Convolution (CPDC), Angular Pixel Difference Convolution (APDC), and Radial Pixel Difference Convolution (RPDC).

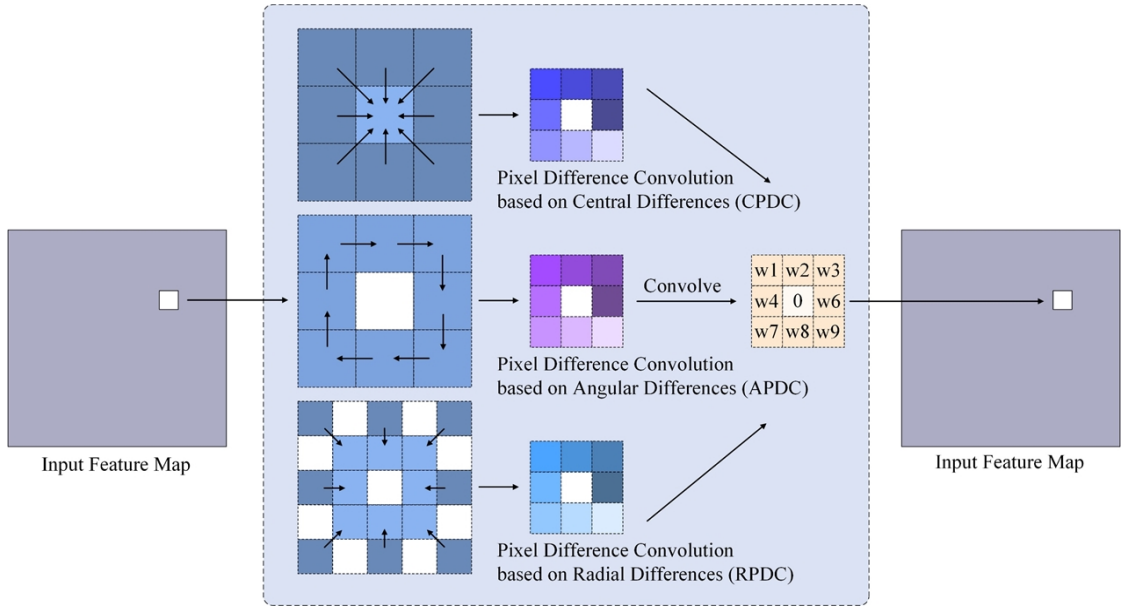

**Fig. 2.** Three types of Pixel Difference Convolution [40].

The proposed detail enhancement module is illustrated in Fig. 3. Inspired by the success of Inception modules [47], we designed an Inception module for underwater image enhancement that includes three groups of parallel Pixel Difference Convolutions (PDCInc). First, a 1×1 convolution is used for preprocessing and controlling the number of feature channels. Subsequently, three different types of PDC are used for further processing, while vanilla convolution is retained. This allows the PDCInc module to capture richer image textures and detail features, enhancing feature diversity and reducing reliance on a single pathway. The features processed in parallel are then concatenated along the channel dimension. After the PDCInc module, another 1×1 convolution is applied for post-processing and channel adjustment. Additionally, the DEM module introduces parallel processing with 3×3 convolutions, which aids in extracting image features at different scales and preserving more shallow features.

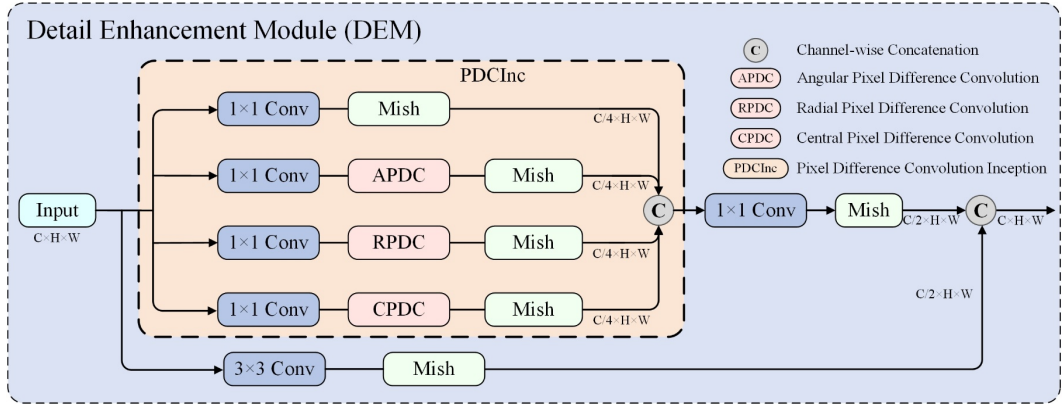

**Fig. 3.** Our proposed detail enhancement module.

### 3.3. Feature fusion module

This paper proposes a feature fusion module for effective information interaction and enhancement between features at different levels, as shown in Fig. 1(b). It takes two inputs, $x_1 \in \mathbb{R}^{C_1 \times H \times W}$ and $x_2 \in \mathbb{R}^{C_2 \times H \times W}$, and first concatenates them along the channel dimension. The resulting features are then fed into a Squeeze-and-Excitation (SE) channel attention mechanism for feature selection along the channel dimension. After that, a 1×1 convolution is applied for processing and channel adjustment, yielding intermediate features $f_c$. Additionally, $x_1$ and $x_2$ are each processed with a 1×1 convolution to adjust their channel numbers, and the resulting outputs are multiplied element-wise with $f_c$ producing two intermediate variables $f_1$ and $f_2$. $f_c$, $f_1$, and $f_2$ can be represented by the following formulas:

$$\begin{aligned} f_c &= \text{Conv}(\text{SE}(\text{Concat}(x_1, x_2))) \\ f_1 &= \text{Conv}(x_1) \otimes f_c \\ f_2 &= \text{Conv}(x_2) \otimes f_c \end{aligned} \quad (2)$$

wehre, Concat denotes the concatenation operation along the channel dimension, SE refers to the channel attention mechanism module, Conv indicates the convolution operation, and $\otimes$ represents element-wise multiplication.

These operations ensure sufficient interaction and enhancement between features from different levels, allowing for precise capture of detail information during the image enhancement process. The various intermediate features are then concatenated along the channel dimension, and the combined features are processed through the SE channel attention mechanism. Following this, a 1×1 convolution is applied to adjust the channel numbers, resulting in the output $y \in \mathbb{R}^{C_3 \times H \times W}$:

$$y = \text{Conv}(\text{SE}(\text{Concat}(f_c, f_1, f_2))) \tag{3}$$

The proposed feature fusion module effectively enhances feature interaction from different input levels through concatenation and multiplication operations, along with the judicious application of the SE channel attention mechanism. This improves the model's efficiency in utilizing multi-level information, ensuring high-fidelity processing of details in the image enhancement task and enhancing the expressiveness of the output features. Additionally, since color degradation in underwater images is channel-related, the introduction of the SE channel attention mechanism aids in improving the color recovery of underwater images.

### 3.4. Loss function

Our proposed PDCFNet utilizes the $l_2$ loss, Structural Similarity (SSIM) loss $l_{SSIM}$, and edge loss $l_{edge}$ [48]. The $l_2$ compares the generated enhanced image $x$ with the reference ground truth $y$ at pixel level, regulating the details and textures of the generated image:

$$l_2 = \| x - y \|_2^2 \tag{4}$$

SSIM loss is used to regulate the similarity between the enhanced image and the reference ground truth:

$$l_{SSIM} = 1 - \text{SSIM}(x, y) \tag{5}$$

where $\text{SSIM}(x, y)$ represents the structural similarity. This is based on image patches and compares the brightness, contrast, and structure between the images. More details can be found in [49]. The edge loss is defined as:

$$l_{edge} = \sqrt{\| \text{Lap}(x) - \text{Lap}(y) \|^2 + \varepsilon^2} \tag{6}$$

where Lap represents the Laplacian operator, and $\varepsilon$ is an infinitesimal quantity used to maintain numerical stability. The edge loss enhances the fidelity and realism of high-frequency details. Finally, the total loss function is given by:

$$l_{total} = l_2 + l_{SSIM} + \lambda l_{edge} \tag{7}$$

where $\lambda$ is set to 0.05 according to [48].

## 4. Experiment and analysis

In this section, we briefly describe the experimental setup, including implementation details, datasets, compared methods and evaluation metrics. Next, we conduct detailed quantitative and qualitative analyses of PDCFNet on three representative datasets and compare it with state-of-the-art models. Finally, we perform histogram comparison, white balance test, and ablation study.

### 4.1. Setup

#### 4.1.1. Implementation details

The experiments were conducted on a windows PC equipped with an Intel Core i7-14700K CPU, 32GB of RAM, and an NVIDIA GeForce RTX 4090D GPU with 24GB of VRAM. The project utilized the PyTorch framework and involved 200 training iterations. The batch size was set to 1. Our model was trained using the ADAM optimizer with a learning rate of $lr = 2 \times 10^{-5}$. We conducted comprehensive qualitative and quantitative evaluations on three representative public datasets.

#### 4.1.2. Datasets

We selected three representative datasets, UIEB [50], EUVP [51], and U45 [52], to train and test our model. These datasets are divided into two categories: 1) Full-reference datasets: 890 image pairs from UIEB and 1200 image pairs from EUVP. 2) Non-reference datasets: 60 challenging images (Challenging 60) from UIEB and 45 images from U45. To facilitate the training and testing of different methods, the images in all three datasets were uniformly resized to 256×256. The structure and division of the datasets are shown in Table 1.

Table 1 The amount and distribution of the underwater datasets.

| Datasets | Trian | | Test | |
| --- | --- | --- | --- | --- |
| | Paired | Unpaired | Paired | Unpaired |
| UIEB | 790 | - | 100 | 60 |
| EUVP | 1000 | - | 200 | - |
| U45 | - | - | - | 45 |

#### 4.1.3. Compared methods and evaluation metrics

Our method was compared with the state-of-the-art methods. The traditional methods include: BTLM [8], UNTV [53], MLLE [54], HLRP [13], PCDE [55],

WWPF [56], HFM [57]. Deep learning-based methods: Ucolor [17], PUIE [16], UUUIR [58], USUIR [59], U-shape [34], UDAformer [33], DICAM [60]. To ensure a fair comparison, all parameter settings in the comparison models follow the configurations provided in the original papers, except for necessary modifications related to image size.

The full reference evaluation metrics and the on-reference evaluation metric were used to quantitatively evaluate and analyze the generated enhanced images. Full-reference evaluation metrics include MSE, PSNR [61], SSIM [49], and PCQI [62]. MSE compares the differences between the generated image and the reference pixel by pixel. PSNR is a metric based on MSE that represents the difference between the generated image and the GT. SSIM compares images at the block level based on brightness, contrast, and structure. PCQI evaluates image quality from three levels: average intensity, signal intensity and signal structure. A smaller MSE value indicates better image quality, while higher PSNR, SSIM and PCQI values indicate better image quality.

No-reference evaluation metrics mainly include UICM, UISM, UIConM, UIQM [63] and UCIQE [64]. UICM, UISM, and UIConM represent the color index, sharpness index, and contrast index, respectively. UIQM is a linear combination of these three indices. UCIQE provides a comprehensive evaluation of underwater images based on color concentration, saturation, and contrast. For all these no-reference evaluation metrics, higher values indicate better image quality.

**4.2. Quantitative evaluation**

The performance of different methods on the UIEB dataset is shown in Table 2. Our proposed PDCFNet achieved the best results in terms of MSE, PSNR, and SSIM. It ranked third in PCQI and UIQM metrics.

**Table 2** The performance of different methods on the UIEB dataset. The best/second-best performance are highlighted in bold/bold italics. This applies to the tables in the following sections unless otherwise specified.

| Methods | MSE($\times 10^3$)↓ | PSNR(dB)↑ | SSIM↑ | PCQI↑ | UIQM↑ | UCIQE↑ |
|---|---|---|---|---|---|---|
| BLTM TB2020 | 1.30±1.17 | 18.58±3.93 | 0.77±0.09 | 0.79±0.18 | 2.30±0.70 | 0.62 |
| UNTV TCSVT2021 | 1.67±0.95 | 16.64±2.67 | 0.54±0.09 | 0.65±0.15 | 1.94±0.62 | 0.63 |
| MLLE TIP2022 | 1.18±1.14 | 18.64±3.11 | 0.73±0.09 | **1.06±0.14** | 2.37±0.53 | 0.61 |
| HLRP TIP2022 | 1.26±1.37 | 19.17±4.19 | 0.71±0.17 | 0.89±0.16 | 2.64±0.84 | **0.66** |
| PCDE SPL2023 | 1.94±1.72 | 16.46±3.09 | 0.65±0.16 | 0.92±0.14 | 2.12±0.73 | 0.62 |
| WWPF TCSVT2023 | 1.10±0.82 | 18.80±3.07 | 0.78±0.08 | *1.02±0.13* | 2.64±0.45 | 0.61 |

| Methods | MSE(×10³)↓ | PSNR(dB)↑ | SSIM↑ | PCQI↑ | UIQM↑ | UCIQE↑ |
|---|---|---|---|---|---|---|
| HFM EAAI2024 | 0.77±0.54 | 20.26±3.02 | 0.79±0.08 | 0.91±0.15 | 2.86±0.39 | 0.63 |
| Ucolor TIP2021 | 0.36±0.35 | 24.18±3.71 | 0.89±0.07 | 0.82±0.14 | **3.19±0.38** | 0.60 |
| PUIE ECCV2022 | 0.38±0.35 | 24.12±4.34 | 0.84±0.09 | 0.71±0.15 | 3.09±0.38 | 0.58 |
| UUUIR ICASSP2022 | 1.20±1.42 | 19.06±3.70 | 0.80±0.10 | 0.86±0.17 | 2.59±0.67 | 0.62 |
| USUIR AAAI2022 | 0.37±0.29 | 23.88±3.91 | 0.85±0.07 | 0.85±0.14 | 3.11±0.30 | *0.65* |
| U-shape TIP2023 | 0.51±0.41 | 22.05±2.98 | 0.78±0.09 | 0.62±0.14 | *3.15±0.36* | 0.58 |
| UDAformer CG2023 | *0.24±0.27* | *26.88±5.11* | *0.92±0.07* | 0.92±0.14 | 3.04±0.41 | 0.62 |
| DICAM NC2024 | 0.30±0.46 | 26.73±5.91 | 0.91±0.08 | 0.85±0.14 | 3.07±0.46 | 0.60 |
| PDCFNet (ours) | **0.22±0.22** | **27.37±5.27** | **0.92±0.06** | 0.92±0.13 | 3.12±0.42 | 0.62 |

The performance of different methods on the EUVP dataset is shown in Table 3. Since PUIE requires four different labels for training and EUVP lacks the corresponding labels, we did not compare PUIE on this dataset. Similar to the UIEB dataset, our proposed PDCFNet achieved the best results in terms of MSE, PSNR, and SSIM metrics. Although PDCFNet did not rank first or second in the other three metrics, its performance was still commendable.

**Table 3** The performance of different methods on EUVP dataset.

| Methods | MSE(×10³)↓ | PSNR(dB)↑ | SSIM↑ | PCQI↑ | UIQM↑ | UCIQE↑ |
|---|---|---|---|---|---|---|
| BLTM TB2020 | 0.88±0.40 | 19.14±1.98 | 0.64±0.07 | 0.74±0.12 | 1.86±0.50 | **0.64** |
| UNTV TCSVT2021 | 1.10±0.62 | 18.25±2.14 | 0.54±0.09 | 0.72±0.11 | 1.91±0.63 | 0.62 |
| MLLE TIP2022 | 1.51±0.75 | 16.83±2.05 | 0.58±0.07 | **0.83±0.12** | 2.01±0.53 | 0.60 |
| HLRP TIP2022 | 1.69±1.37 | 17.25±3.60 | 0.58±0.13 | 0.73±0.13 | 2.47±0.81 | **0.64** |
| PCDE SPL2023 | 1.78±0.96 | 16.09±1.92 | 0.55±0.10 | 0.80±0.12 | 1.75±0.76 | 0.61 |
| WWPF TCSVT2023 | 1.41±0.97 | 17.29±2.25 | 0.61±0.08 | **0.83±0.12** | 2.40±0.40 | 0.61 |
| HFM EAAI2024 | 0.73±0.50 | 20.25±2.47 | 0.70±0.08 | 0.75±0.12 | 3.00±0.33 | 0.62 |
| Ucolor TIP2021 | 0.26±0.26 | 25.23±2.95 | 0.79±0.08 | 0.64±0.10 | *3.21±0.35* | 0.57 |
| UUUIR ICASSP2022 | 1.63±1.23 | 17.13±3.17 | 0.63±0.10 | 0.73±0.13 | 2.42±0.49 | 0.63 |
| USUIR AAAI2022 | 0.59±0.55 | 21.49±2.90 | 0.69±0.07 | 0.72±0.12 | 2.68±0.23 | 0.63 |
| U-shape TIP2023 | 0.77±0.82 | 21.44±4.49 | 0.73±0.11 | 0.57±0.10 | **3.23±0.36** | 0.55 |
| UDAformer CG2023 | *0.21±0.19* | 26.20±3.13 | 0.82±0.06 | 0.69±0.12 | 3.16±0.40 | 0.58 |
| DICAM NC2024 | 0.21±0.20 | *26.35±3.23* | *0.83±0.06* | 0.69±0.11 | 3.15±0.41 | 0.58 |
| PDCFNet (ours) | **0.19±0.19** | **26.67±3.04** | **0.84±0.06** | 0.73±0.11 | 3.11±0.44 | 0.59 |

To evaluate the robustness of the proposed method, we conducted tests on the no-reference datasets Challenging60 and U45. Due to the lack of ground truth for training, we tested the model trained on UIEB on these datasets, with results presented in Table 4. Since our method leverages pixel difference convolution to enhance image detail, PDCFNet achieved the best performance in UISM, which represents image clarity. For the other metrics, PDCFNet also demonstrated relatively good performance.

Table 4 The performance results of different methods on Challenging60 and U45 datasets.

| Methods | Challenging60 | | | | | U45 | | | | |
|---|---|---|---|---|---|---|---|---|---|---|
| | UICM↑ | UISM↑ | UIConM↑ | UIQM↑ | UCIQE↑ | UICM↑ | UISM↑ | UIConM↑ | UIQM↑ | UCIQE↑ |
| BLTM TB2020 | 6.24±3.27 | 3.58±1.67 | 0.19±0.09 | 1.91±0.53 | 0.59 | *8.15±3.70* | 5.54±1.28 | 0.18±0.13 | 2.52±0.75 | 0.60 |
| UNTV TCSVT2021 | 4.90±2.23 | 3.11±1.67 | 0.16±0.09 | 1.62±0.68 | 0.59 | 6.32±2.28 | 4.48±1.46 | 0.10±0.07 | 1.84±0.62 | 0.62 |
| MLLE TIP2022 | 3.44±1.47 | 4.23±1.43 | 0.17±0.08 | 1.96±0.55 | 0.59 | 4.23±1.55 | 6.16±1.24 | 0.18±0.06 | 2.60±0.50 | 0.59 |
| HLRP TIP2022 | **8.50±3.75** | 3.35±2.51 | 0.18±0.07 | 1.87±0.93 | **0.64** | 8.07±3.70 | 6.32±1.82 | 0.24±0.08 | 2.96±0.73 | **0.64** |
| PCDE SPL2023 | 4.89±2.27 | 3.38±1.66 | 0.17±0.08 | 1.74±0.71 | 0.60 | 5.09±2.33 | 4.75±2.52 | 0.18±0.08 | 2.18±0.99 | 0.61 |
| WWPF TCSVT2023 | 4.03±1.67 | 5.51±1.62 | 0.20±0.07 | 2.47±0.51 | 0.59 | 4.51±1.67 | 6.93±0.81 | 0.21±0.06 | 2.93±0.35 | 0.60 |
| HFM EAAI2024 | 4.40±2.08 | 4.57±2.11 | 0.23±0.03 | 2.31±0.58 | 0.59 | 5.82±2.40 | 7.26±1.03 | 0.23±0.04 | 3.12±0.27 | 0.61 |
| Ucolor TIP2021 | 4.46±2.00 | 5.58±1.92 | 0.30±0.03 | **2.85±0.60** | 0.56 | 5.55±2.01 | 7.40±0.85 | 0.29±0.03 | **3.38±0.28** | 0.59 |
| PUIE ECCV2022 | 3.66±1.69 | 4.94±1.78 | **0.32±0.03** | 2.71±0.54 | 0.55 | 4.44±1.61 | 6.68±0.90 | **0.33±0.02** | 3.26±0.30 | 0.57 |
| UUUIR ICASSP2022 | 6.72±2.87 | 4.67±1.74 | 0.24±0.08 | 2.43±0.62 | 0.58 | 7.46±3.47 | 6.18±1.30 | 0.22±0.12 | 2.81±0.73 | 0.60 |
| USUIR AAAI2022 | *7.54±2.42* | 5.50±1.69 | 0.28±0.03 | *2.83±0.48* | *0.63* | 8.53±2.71 | 7.24±0.92 | 0.26±0.03 | 3.31±0.18 | *0.63* |
| U-shape TIP2023 | 3.91±1.86 | 5.31±1.73 | *0.31±0.03* | 2.80±0.55 | 0.55 | 4.97±2.01 | 6.76±0.96 | *0.32±0.03* | 3.27±0.29 | 0.57 |
| UDAformer CG2023 | 5.28±2.43 | *5.63±1.93* | 0.27±0.05 | 2.79±0.55 | 0.59 | 6.64±2.32 | 7.35±0.81 | 0.24±0.05 | 3.22±0.25 | 0.61 |
| DICAM NC2024 | 4.81±2.30 | 5.07±2.13 | 0.29±0.05 | 2.67±0.64 | 0.58 | 6.13±2.11 | *7.42±0.98* | 0.27±0.05 | 3.31±0.32 | 0.60 |
| PDCFNet (ours) | 5.16±2.30 | **5.67±1.99** | 0.27±0.04 | 2.78±0.60 | 0.59 | 6.53±2.49 | **7.65±0.77** | 0.25±0.05 | *3.35±0.22* | 0.61 |

Overall quantitative analysis shows that PDCFNet performs exceptionally well on full-reference metrics, achieving the best results in MSE, PSNR, and SSIM. Its performance on no-reference metrics is also commendable, demonstrating the effectiveness and robustness of PDCFNet across different datasets. Additionally, our model has 1.82M parameters and 172.77 GMACs, with an inference time of 23ms, equivalent to 43.48 FPS.

**4.3. Qualitative evaluation**

Different underwater scenarios present unique challenges for image enhancement. To validate the performance of our proposed method in various scenarios, we conducted qualitative analyses across different scenarios. Based on the characteristics of underwater environments, we selected four representative and challenging scenarios: color deviation, blurriness, scattering, and darkness. The performances of different methods are illustrated in Figs. 4 to 7.

Light attenuation underwater leads to color distortion, with red light attenuating the fastest and blue-green light attenuating more slowly, resulting in underwater images appearing predominantly blue-green, as shown in Fig. 4. Among traditional methods, MLLE and HFM performed well. In the deep learning approaches, except for UUUIR and USUIR, other methods effectively corrected the color of underwater images.

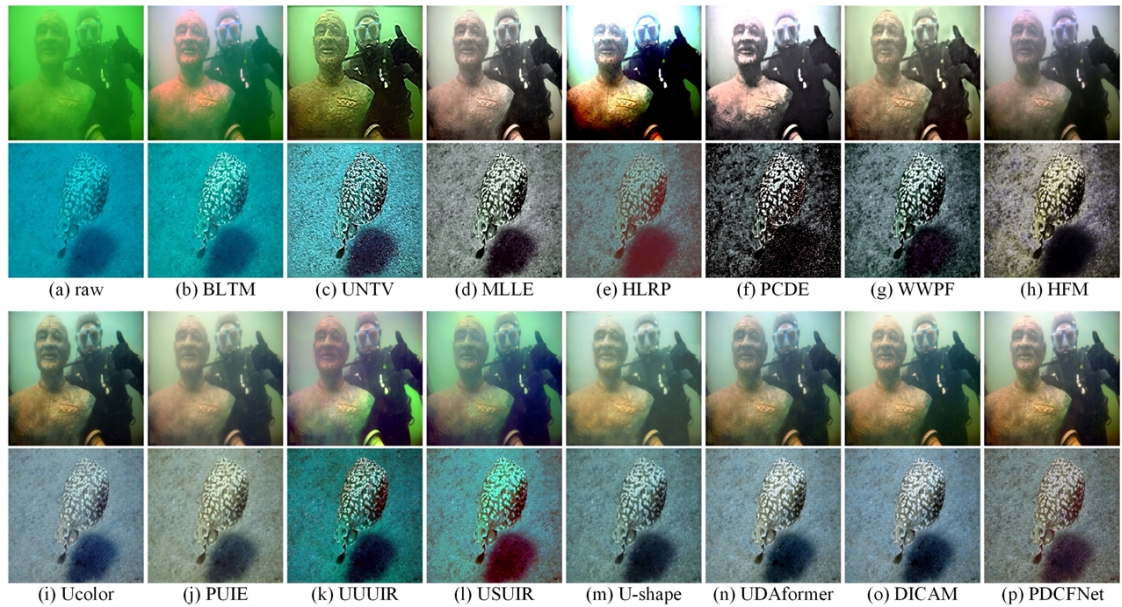

Fig. 4. Comparison of color deviation scenarios from U45 dataset.

When light passes through uneven media such as marine organisms, sediment particles, bubbles, or water masses, its propagation path undergoes complex changes, altering the spatial distribution of light and leading to further scattering. In severe cases, this can result in the Tyndall effect. In such situations, most physical models fail, presenting significant challenges for UIE. As shown in Fig. 5, UUUIR and traditional methods do not effectively enhance images in scattering scenarios. Other deep learning methods exhibit varying degrees of over-enhancement. The proposed PDCFNet performs relatively well in this scenario, avoiding excessive enhancement of the images.

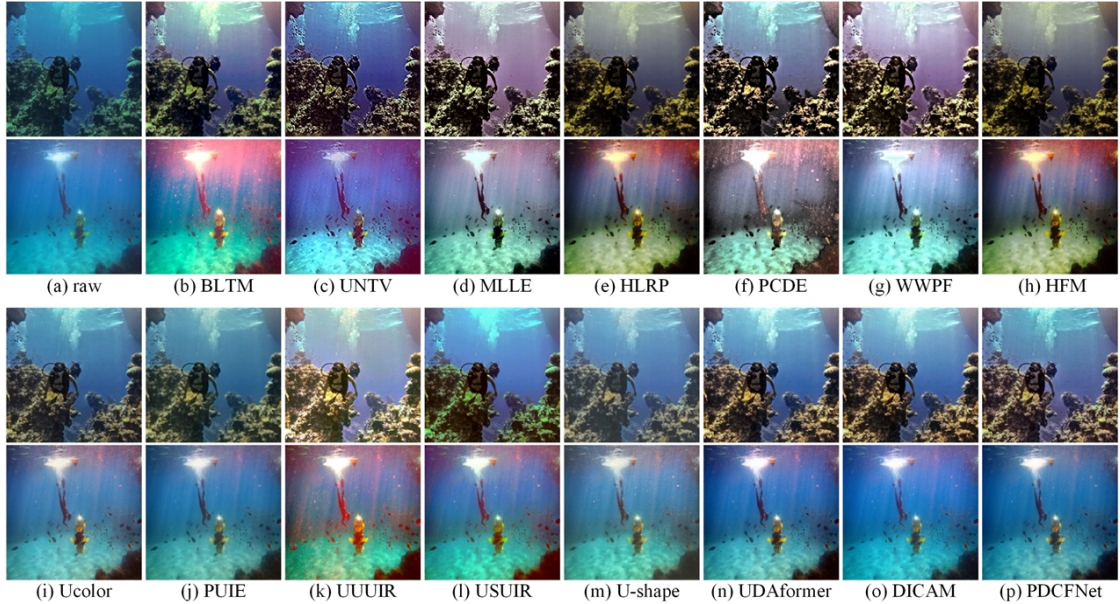

**Fig. 5.** Comparison of scattering scenarios from UIEB.

Due to light attenuation and the impact of suspended particles, underwater images often exhibit low contrast and blurriness. As shown in Fig. 6, traditional methods such as UNTV, MLLE, WWPF, and HFM effectively reduce blurriness. Among the deep learning methods, USUIR, UDAformer, DICAM, and the proposed PDCFNet also perform well in removing blurriness and enhancing the details in the images.

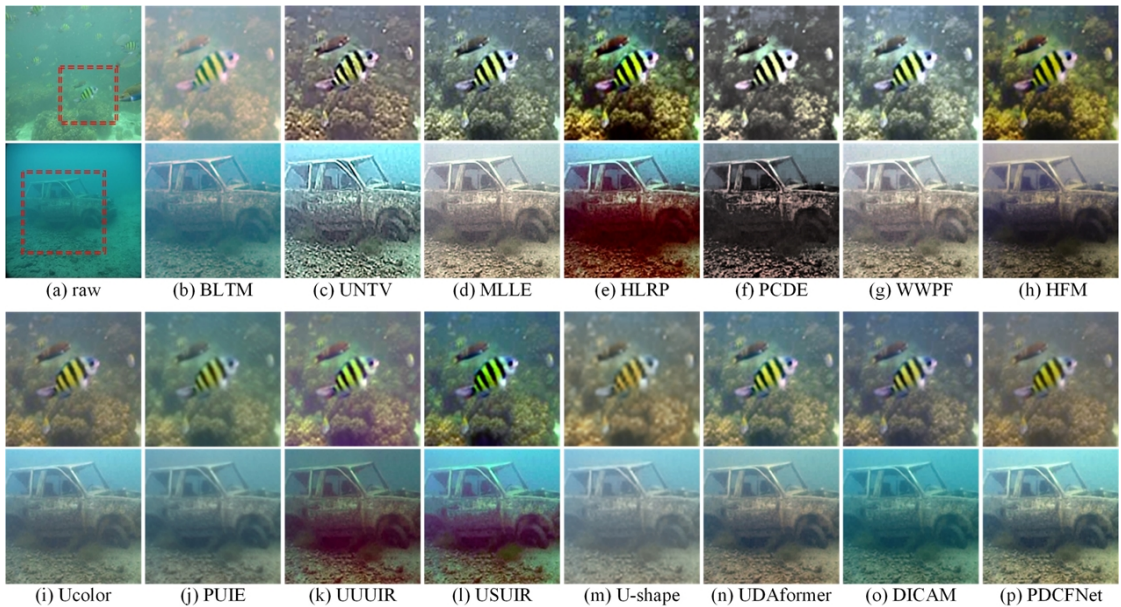

**Fig. 6.** Comparison of blurred scenarios from UIEB.

As depth increases, light gradually diminishes. In dark scenarios, image contrast decreases significantly, leading to substantial detail and color loss. Additionally, artificial light sources can create localized bright spots, complicating the balance of global illumination. These challenges make UIE more complex under dark conditions. As shown in Fig. 7, most traditional methods, except for HFM, struggle with inadequate enhancement in dark areas and excessive color boosting, introducing unrealistic reds. Among the deep learning methods, UUUIR and UDAformer exhibit some over-enhancement. The proposed PDCFNet achieves a better balance in global illumination, underwater color recovery, and detail enhancement in dark regions.

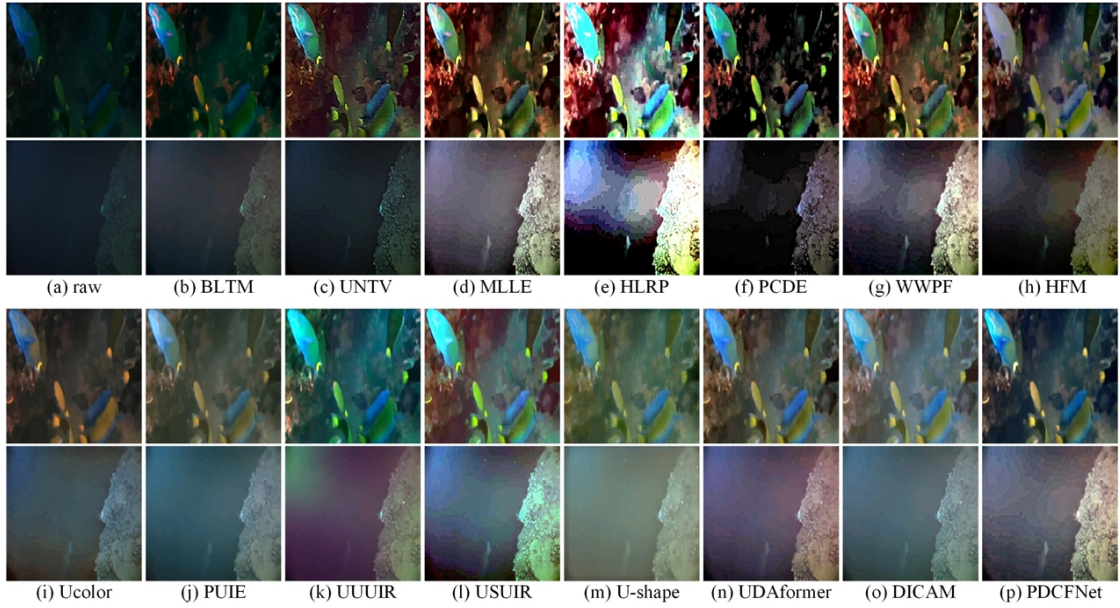

**Fig. 7.** Comparison of dark scenarios from Challenging60.

Our proposed PDCFNet performs well in four underwater degradation scenarios, demonstrating its effectiveness and robustness.

**4.4. Histogram Comparisons and white balance test**

To comprehensively demonstrate the effectiveness of the proposed PDCFNet in color restoration, we conducted histogram comparison and white balance test. Red light attenuates significantly underwater, leading to lower corresponding pixel values. This is reflected in the histogram, where the red curve skews to the left, as shown in Fig. 8. Traditional methods and UUUIR tend to spread the RGB pixel values towards the minimum and maximum ends, failing to effectively restore the color diversity of underwater images. In contrast, PDCFNet achieves a more uniform distribution of the RGB curves, demonstrating its ability to suppress the dominant channel and compensate for the attenuated channels, effectively correcting the colors of underwater images.

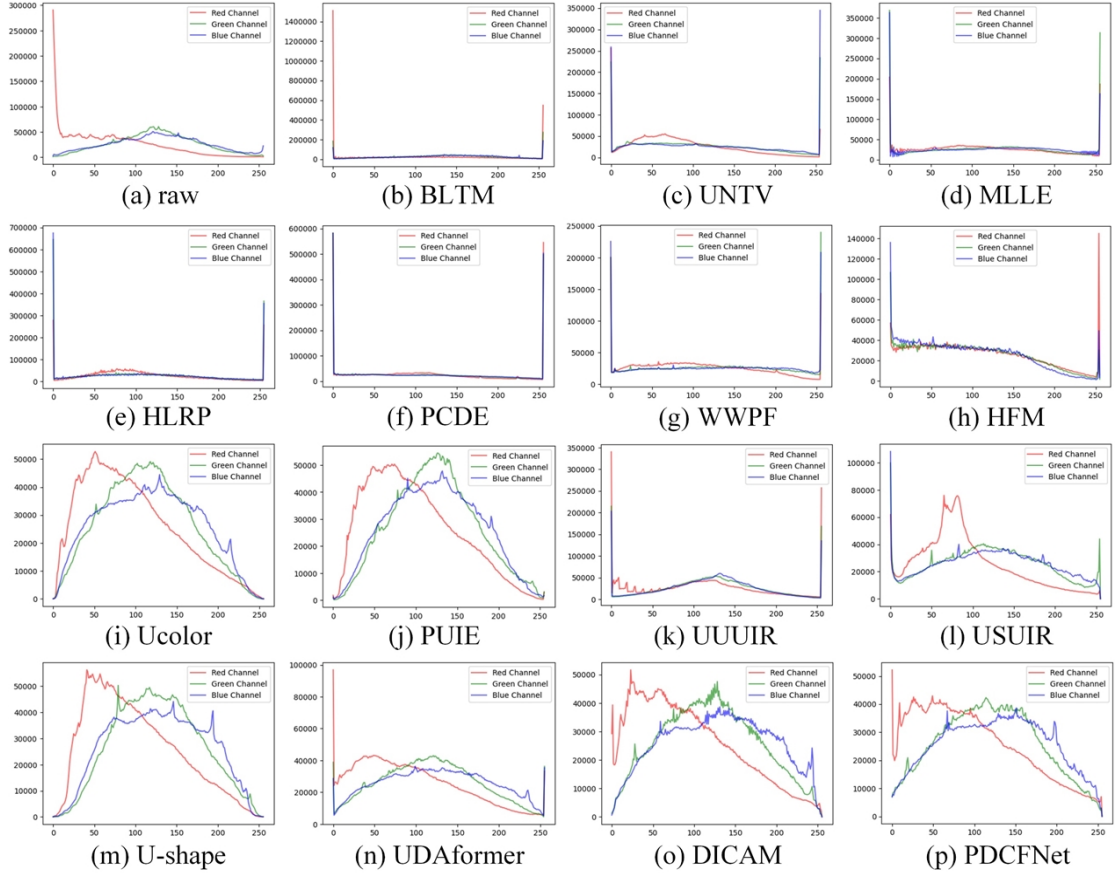

**Fig. 8.** Comparison of RGB histogram distributions on the UIEB dataset. The vertical axis represents frequency, and the horizontal axis represents pixel values.

To further visually demonstrate the performance of PDCFNet in color restoration, we conducted white balance test. This testing was performed on the Color-Checker7 dataset [65], which consists of images captured by divers holding a standard color card in a swimming pool. The results are shown in Fig. 9, with a standard color card included in the top left corner of the raw image. Aside from WWPF, which performed relatively well, other traditional methods exhibited either insufficient or excessive color correction. Among deep learning methods, UUUIR showed a significant deficiency in color restoration, while other approaches, although better at correcting colors, fell short in enhancing visibility. Our proposed PDCFNet not only restores colors effectively but also improves the visibility of underwater images.

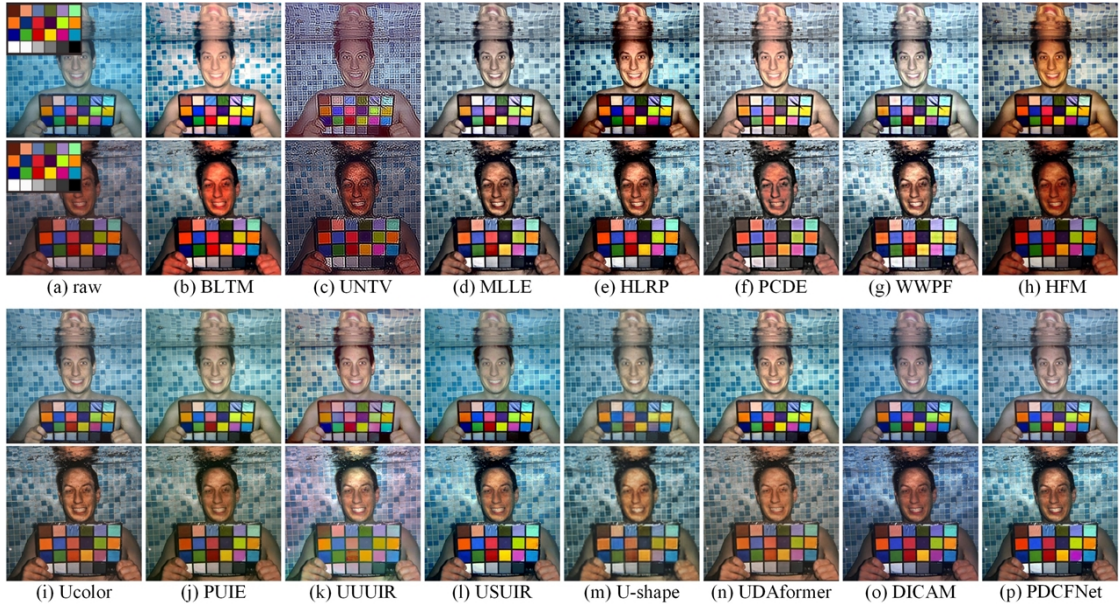

**Fig. 9.** Results of white balance test.

### 4.5. Ablation study

To better understand the proposed method, we conducted ablation study on the UIEB dataset focusing on PDC and the loss function. First, we removed the PDCs from the detail enhancement module, replacing them with a 3×3 convolutions. Secondly, we excluded different components from the loss function. The experimental results are shown in Table 5. After removing PDCs, all metrics declined. However, due to our robust network design, the performance of the network without PDCs remained satisfactory. In the loss function, excluding certain losses led to improvements in some metrics. For example, removing the SSIM loss resulted in better performance in PCQI and UIQM, while other metrics decreased. Overall, maintaining all losses proved to be the best option.

**Table 5** Ablation study on the UIEB dataset. w/o means removing specific components.

| Methods | MSE(×10³)↓ | PSNR(dB)↑ | SSIM(×10²)↑ | PCQI↑ | UIQM↑ | UCIQE↑ |
|---|---|---|---|---|---|---|
| PDCFNet | 0.22±0.22 | **27.37±5.27** | **92.02±5.92** | 0.92±0.13 | 3.12±0.42 | **0.62** |
| w/o PDC | 0.23±0.27 | 26.96±5.26 | 91.27±6.56 | 0.90±0.14 | 3.11±0.42 | 0.61 |
| w/o $l_2$ | **0.22±0.21** | 26.66±4.56 | 91.57±6.35 | 0.91±0.14 | 3.13±0.39 | **0.62** |
| w/o edge | 0.24±0.24 | 26.65±4.98 | 91.64±6.12 | 0.93±0.14 | 3.13±0.40 | **0.62** |
| w/o SSIM | 0.23±0.23 | 26.62±4.76 | 91.30±5.61 | **0.95±0.12** | **3.19±0.38** | 0.62 |

### 5. Conclusion

This paper proposes a single underwater image enhancement network, PDCFNet, based on pixel difference convolution and cross-level feature fusion. Specifically, we designed a detail enhancement module by introducing pixel difference convolution to capture high-frequency features, achieving better detail and texture enhancement. The designed cross-level feature fusion module performs operations such as concatenation and multiplication on features from different layers, ensuring sufficient interaction and enhancement between them. Extensive experiments demonstrate that PDCFNet achieves state-of-the-art performance in both quantitative metrics and qualitative analysis, effectively enhancing underwater images in various scenarios. Despite PDCFNet's impressive performance in UIE, there remains considerable room for exploration in future research.

## CRediT authorship contribution statement

**Song Zhang:** Writing-original draft, Visualization, Validation, Software, Methodology, Conceptualization. **Daoliang Li:** Writing–review & editing, Funding acquisition. **Ran Zhao:** Writing–review & editing, Supervision, Methodology, Funding acquisition, Conceptualization.

## Declaration of competing interest

The authors declare that they have no known competing financial interests or personal relationships that could have appeared to influence the work reported in this paper.

## Data availability

Data will be made available on request.

## Acknowledgments

This paper was supported by The National Natural Science Foundation of China (NO.32273188) and Key Research and Development Plan of the Ministry of Science and Technology (NO.2022YFD2001700).